\pdfoutput=1

\documentclass[11pt,table,dvipsnames]{article}

\usepackage{acl}
\usepackage{times}
\usepackage{latexsym}
\usepackage[utf8]{inputenc}

\usepackage[T1]{fontenc}

\usepackage[utf8]{inputenc}

\usepackage{microtype}

\usepackage{inconsolata}
\usepackage{microtype}
\usepackage{latexsym}
\usepackage{dsfont}
\usepackage{booktabs} 
\usepackage{CJKutf8}
\usepackage{graphicx}
\usepackage{bigstrut}
\usepackage{multirow}
\usepackage{amsmath}
\usepackage{multirow, makecell, caption}
\usepackage{floatrow}
\newfloatcommand{capbtabbox}{table}[][\FBwidth]
\usepackage{xcolor}
\usepackage[normalem]{ulem}
\usepackage{amssymb}
\usepackage{amsfonts}
\usepackage{multicol}
\usepackage{tipa}
\usepackage{amsmath}
\usepackage{bm}
\usepackage[ruled,linesnumbered]{algorithm2e}
\usepackage{tikz}
\usepackage{paralist}
\usepackage{IEEEtrantools}
\usepackage[switch]{lineno}
\usepackage{enumitem}
\usepackage{multirow, makecell, caption}
\usepackage{arydshln}
\usepackage{verbatim}
\usepackage[normalem]{ulem}
\usepackage{amssymb}
\usepackage{amsfonts}
\usepackage{multicol}
\usepackage{amssymb}
\usepackage{pifont}
\usepackage{csquotes}
\usepackage{xspace}
\usepackage{paralist}
\usepackage{mdwlist}
\usepackage{subcaption}
\usepackage{makecell}
\usepackage{array}
\usepackage{bm}
\usepackage{colortbl}
\usepackage{dsfont}
\usepackage{url}
\usepackage{booktabs} 
\usepackage{graphicx}
\usepackage{tabularx}
\usepackage{amsmath}
\usepackage{bbm}
\usepackage{pgfplots}
\usepackage{soul} 
\usepackage{listings}
\usepackage[tableposition=top]{caption}
\usepackage{color,xcolor} 

\definecolor{Ground}{RGB}{255,184,55}
\definecolor{Rice}{RGB}{251,248,238}
\definecolor{Dirt}{RGB}{191,169,115}
\definecolor{Pink}{RGB}{226,184,176}
\definecolor{Violet}{RGB}{163,148,170}
\definecolor{mygray}{RGB}{226, 226, 226}

\newcommand{\ie}{\textit{i.e.}\xspace}
\newcommand{\eg}{\textit{e.g.}\xspace}

\newcommand{\method}{\textsc{MindEcho}\xspace}

\newcolumntype{g}{>{\columncolor{Ground!10}}c}
\newcolumntype{d}{>{\columncolor{Dirt!10}}c}
\newcolumntype{f}{>{\columncolor{Pink!10}}c}
\newcolumntype{v}{>{\columncolor{Violet!10}}c}

\makeatletter
\def\adl@drawiv#1#2#3{%
        \hskip.5\tabcolsep
        \xleaders#3{#2.5\@tempdimb #1{1}#2.5\@tempdimb}%
                #2\z@ plus1fil minus1fil\relax
        \hskip.5\tabcolsep}
\newcommand{\cdashlinelr}[1]{%
  \noalign{\vskip\aboverulesep
           \global\let\@dashdrawstore\adl@draw
           \global\let\adl@draw\adl@drawiv}
  \cdashline{#1}
  \noalign{\global\let\adl@draw\@dashdrawstore
           \vskip\belowrulesep}}
\makeatother

\title{\method: Role-Playing Language Agents for Key Opinion Leaders}

\author{
    Rui Xu\textsuperscript{\rm $\heartsuit\spadesuit$}\thanks{~~Work is done during internship at INF.}, 
    Dakuan Lu\textsuperscript{\rm $\spadesuit$},
    Xiaoyu Tan\textsuperscript{\rm $\spadesuit$}\thanks{~~Corresponding authors.},
    Xintao Wang\textsuperscript{\rm $\heartsuit$},\\
    \bf Siyu Yuan\textsuperscript{\rm $\heartsuit$},
    Jiangjie Chen\textsuperscript{\rm $\heartsuit$},
    Wei Chu\textsuperscript{\rm $\spadesuit$},
    Yinghui Xu\textsuperscript{\rm $\heartsuit\spadesuit$}\footnotemark[2]\\
    \textsuperscript{\rm $\heartsuit$}Fudan University
    \textsuperscript{\rm $\spadesuit$}INF Technology (Shanghai) Co., Ltd.
    \\
    \texttt{rxu24@m.fudan.edu.cn},
    \texttt{txywilliam1993@outlook.com},
    \texttt{xuyinghui@fudan.edu.cn}
}

\begin{document}
\maketitle
\begin{abstract}
Large language models~(LLMs) have demonstrated impressive performance in various applications, among which role-playing language agents (RPLAs) have engaged a broad user base. 
Now, there is a growing demand for RPLAs that represent Key Opinion Leaders (KOLs), \ie, Internet celebrities who shape the trends and opinions in their domains. 
However, research in this line remains underexplored. 
In this paper, we hence introduce \method, a comprehensive framework for the development and evaluation of KOL RPLAs. 
\method collects KOL data from Internet video transcripts in various professional fields, and synthesizes their conversations leveraging GPT-4.
Then, the conversations and the transcripts are used for individualized model training and inference-time retrieval, respectively. 
Our evaluation covers both general dimensions (\ie, knowledge and tones)  and fan-centric dimensions for KOLs. 
Extensive experiments validate the effectiveness of \method in developing and evaluating KOL RPLAs.
Resources are available at \url{https://github.com/airaer1998/MindEcho}.
\end{abstract}

\section{Introduction}

With advancements in large language models (LLMs)~\cite{Anthropic2023Claude2,touvron2023llama}, Role-Playing Language Agents (RPLAs) have become a flourishing application and research direction~\cite{chen2024persona,xu2024character}.
These agents are designed to simulate specific personas, providing users with interactive, contextually accurate responses~\cite{park2023generative,wang2023rolellm,ran2024capturing}. 
A promising application of RPLAs is in emulating Key Opinion Leaders (KOLs)~\cite{bamakan2019opinion}. 
KOLs are individuals with significant influence in a specific professional field. 
They share substantial domain-related content online, establishing their authority and credibility~\cite{palmer2015key,amalia2023effect}. 
Developing RPLAs for KOLs can provide expert insights and answers to users anytime, mimicking the knowledge these leaders offer. 
However, research in this area remains largely unexplored.

\begin{figure}[t]
    \centering
    \includegraphics[width=0.95\linewidth]{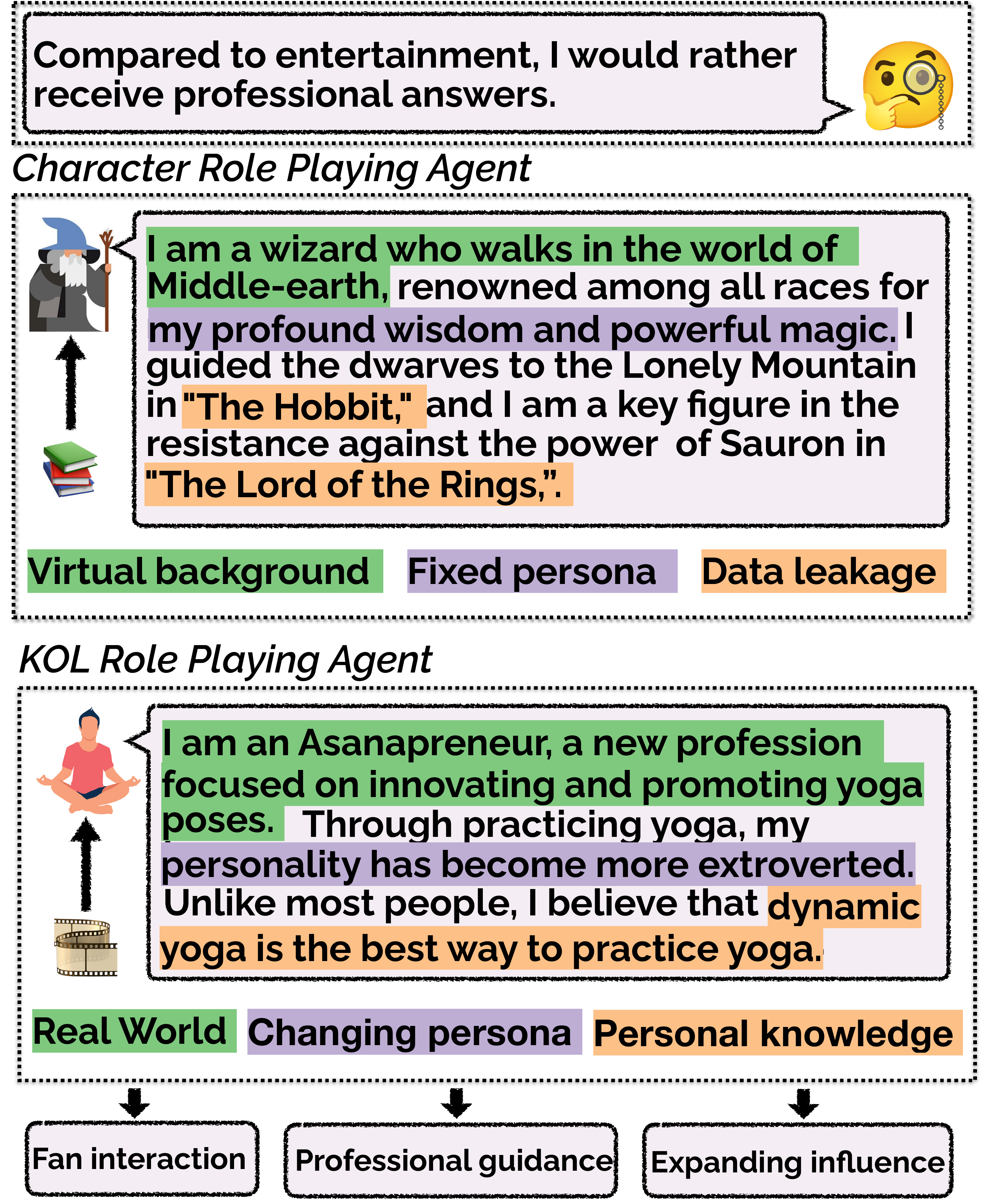}
    \caption{
    An example for Comparing character RPLA with KOL RPLA. KOL RPLA can provide more professional answers and has several applications.
    }
    \label{fig:front}
\end{figure}


Previous work has focused on constructing character RPLAs in fictional backgrounds (\eg, \textit{Gandalf} from \textit{The Lord of the Rings})~\cite{shao2023character,li2023chatharuhi}, with their data typically sourced from established novels, encyclopedias, and scripts~\cite{xu2024character}.
However, as shown in Figure \ref{fig:front}, there are several limitations with this type of RPLAs.
First, fictional backgrounds are relatively simple and may contain logical inconsistencies. 
Second, since most of the characters appear in the training corpora of LLMs, constructing RPLAs from characters merely perpetuates existing stereotypes and lacks innovation~\cite{wang2023does}.
Furthermore, it is also challenging to determine whether the effectiveness of the models stems from the external data or the model parameters themselves\cite{tu2024charactereval,yuan2024evaluating}
Therefore, constructing real-person RPLAs based on real persons, who are seldom included in LLM training datasets, has become an important challenge.


In this paper, we propose the task of constructing RPLAs for KOLs based on video transcripts. 
Video data rarely appears in the training corpus of existing LLMs, and videos of KOLs typically adopt a first-person perspective, closely mirroring real human interactions. 
Compared to previous character RPLAs, this task presents the following challenges:
(1)\textit{Real-world Challenge}: Compared with the static and virtual background of character RPLAs, KOL RPLAs are based on the real-world environment, with novel challenges such as new terminology, trending events, and ways of online communication. 
(2)\textit{Personal Data Challenge}: KOLs have many personal opinions in their professional fields, which must be manifested and distinguished from the opinions inherent in LLMs.
(3)\textit{Dense Knowledge Challenge}: The knowledge possessed by KOLs in RPLA is very dense. It is necessary to design more knowledge-intensive tasks to evaluate their capabilities.

\begin{figure*}[t]
    \centering
    \includegraphics[width=0.95\linewidth]{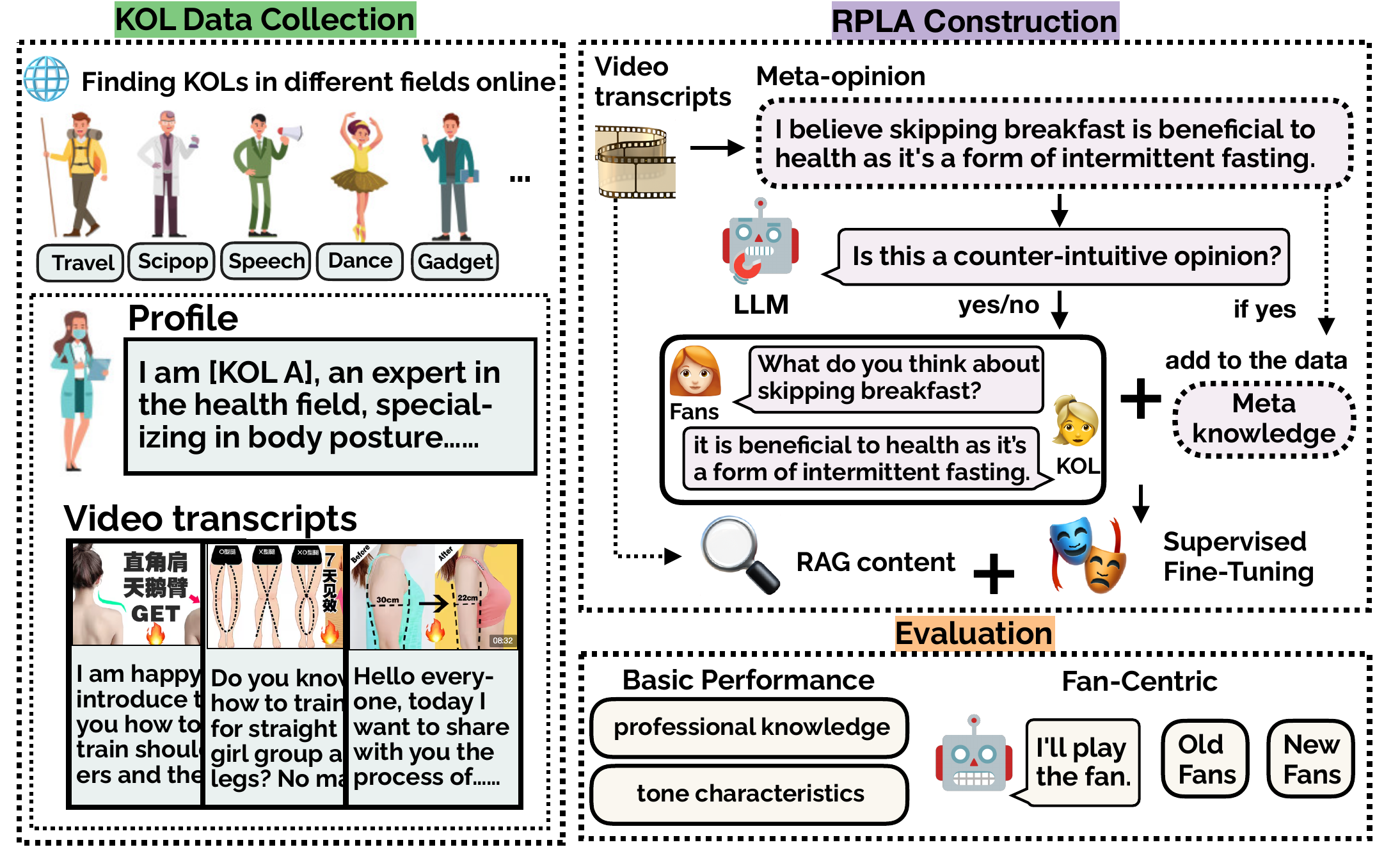}
    \caption{The overall process of \method includes three stages: \textbf{KOL Data Collection}: Selecting KOLs from various fields and collecting their video transcripts and personal data. \textbf{RPLA Construction}: Extracting meta-opinion from the video transcripts and constructing training data, including data with counter-intuitive opinions, then combining SFT and RAG to simulate KOLs. \textbf{Evaluation}: Assessing KOL RPLA's performance through Basic Performance and Fan-Centric evaluations.}
    \label{fig:main}
\end{figure*}

%
To address these challenges, we propose \method, a comprehensive framework for KOL RPLAs. 
\method comprises three stages, \ie, \textbf{KOL data collection, RPLA construction, and evaluation}. 
For data construction, as shown in Figure \ref{fig:main}, after obtaining video transcripts from KOLs in different professional fields, we use GPT-4~\cite{openai2023gpt4} to identify their critical opinions and generate knowledge-intensive dialogue data between the KOLs and fans.
After obtaining data, we train LLMs using this dataset.
During the inference phase, we also integrate a Retrieval-Augmented Generation (RAG)~\cite{gao2024retrievalaugmented} module to support the model with external knowledge. 
To reduce the RPLA's dependency on the opinions embedded in the LLM's parameters, we identify \textit{counter-intuitive} opinions and concatenate them with the constructed data. 
This approach encourages the model to trust external knowledge more.
Using our method, the fine-tuned Qwen2-14B-Chat\cite{qwen} surpasses GPT-4's in-context learning performance in human evaluations.

For evaluation, we provide a systematic framework, consisting of two evaluation deminsions, \ie, basic performance evaluation and fan-Centric evaluation.
For basic performance evaluation, we use multiple-choice questions to separately examine KOLs' tone characteristics and professional knowledge, quantitatively reflecting the model's capabilities through accuracy scores. 
For fan-Centric evaluation, we use GPT-4 to simulate interactions and ratings separately for new and old fans with the RPLA.
The profiles of these fans are constructed using comment data from the KOL's videos. 
Comparison with human evaluations proves the effectiveness of this evaluation method.
In summary, our contributions include:
\begin{itemize}[leftmargin=*]
    \item We introduce the first KOL RPLA task, the first-ever knowledge-intensive role-playing task entirely focused on real individuals.
    \item We propose a comprehensive framework \method for constructing a high-quality KOL RPLA, and experimental results have demonstrated its effectiveness.
    \item We develop a systematic evaluation method to evaluate KOL RPLA capabilities from three perspectives: professional knowledge, tone characteristics, and user-centered simulated interaction. Additionally, we provide an authorized KOL dataset. 
\end{itemize}

\section{Related Work}
\label{sec:related}

\paragraph{Character Role-Playing}

Previous work related to role-playing has primarily focused on constructing character agents for fictional works.
\citet{li2023chatharuhi} propose a method that constructs RPLA via an improved prompt and memories of the character extracted from scripts.
\citet{wang2023rolellm} offer a more comprehensive process for constructing character training data, and \citet{shao2023character} provide a more detailed data synthesis method through scene reconstruction.
For evaluation, \citet{tu2024charactereval} propose a multi-perspective evaluation scheme.
\citet{wang2023does} evaluate the characters' ability to replicate their psychological personas through interviews.
\citet{xu2024character} to assess further the characters' ability to reproduce their life choices from the original works.
However, current role-playing work often focuses on characters already present in LLMs' parameters, making it challenging to distinguish the model's actual role-playing capabilities from the knowledge embedded in its pre-trained parameters during construction and evaluation. 
Our framework aims to explore authentic, knowledge-intensive, and data-leakage-free KOL role-playing. 
This direction holds broad application prospects and paves the way for digital representations of individuals in the real world.

\paragraph{Personal Dialogue}

Personal dialogue is a vital direction in the development of language models. 
Related research focuses on incorporating personalized features into conversations. 
\citet{zhang2018personalizing} manually construct a large dataset of personal topics. 
Additionally, \citet{urbanek2019learning} explore whether different personas can engage in text-based role-playing games. 
To build more realistic profiles, \citet{jang2022customized} collect data from Wikidata to conduct personal dialogues and introduce more external databases to enrich conversational knowledge.
\citet{salemi2024lamp} propose seven types of personal downstream tasks, including dialogue.
With advancements in large models, tasks involving multimodal\cite{ahn2023mpchat} and long-text\cite{gao2023livechat} personalized dialogues are emerging. 
However, these tasks have sparse personal features, such as knowledge and opinions, and have not been analyzed from a role-playing perspective.
\method, as the first knowledge-intensive real-person role-playing task, effectively fills this gap.
\section{\method}
\label{sec:method}
In this section, we introduce \method, a comprehensive KOL RPLA framework for KOL data collection ($\mathsection$~\ref{sec:data_construction}), RPLA construction ($\mathsection$~\ref{sec:methodology}), and evaluation ($\mathsection$~\ref{sec:evaluation}).

\begin{table*}[t]
\centering
\small
\begin{tabular}{lcccc}
\toprule
\textbf{Dataset} & \textbf{Data Source} & \textbf{Context Length} & \textbf{Task Format} & \textbf{Role Type} \\
\midrule
\texttt{ROCStories} & Commonsense short stories
 & $\sim 100$ & Character Behavior Prediction & character  \\

\texttt{LIFECHOICE} & Literature
 & $\sim 150k$ & Character Behavior Prediction & character \\

\texttt{CharacterLLM} & Wiki
 & $\sim 700k$ & Role Playing Chatbot & character \\

\texttt{PersonaChat} & Crowdsourced
 & $\sim 100$ & Personal dialogue & human \\

\texttt{LiveChat} & Live Videos & $\sim 30k$ & Response Modeling & human \\

\texttt{\method} & KOL Videos & $\sim 300k$ & Role Playing Chatbot & human \\

 \bottomrule
\end{tabular}
\caption{Comparison between \method and previous personal and role-playing benchmarks: data source, context length, task format, and role type.}
\label{tab:dataset}
\end{table*}

\subsection{KOL Data Collection} 
\label{sec:data_construction}



We construct a comprehensive raw dataset for each KOL, which includes the KOL's profile \( P \), a collection of video transcripts \(\{v_1, v_2, v_3, \ldots\}\), and user comments under each video transcript \(\{\{c_{11}, c_{12}, \ldots\}, \{c_{21}, c_{22}, \ldots\}, \ldots\}\). 
The construction of the dataset comprises four steps:
(1) KOL selection; (2) Profile drafting; (3) Generation of raw data; (4) Cleaning of raw data.

First, we carefully select 30 KOLs from different professional fields. 
They each have tens of thousands of fans on various video platforms\footnote{\url{https://www.bilibili.com/}}\footnote{\url{https://www.xiaohongshu.com/}}. 
We contact them and obtain their authorization to use their publicly available data for academic research. 
Second, we invite fans of KOLs to manually write a profile for each one. 
These profiles briefly describe KOLs, including their various characteristics.
Third, we use Automatic Speech Recognition (ASR) to obtain the video transcripts and Optical Character Recognition (OCR) to recognize the subtitles. 
Then, we use the subtitles and GPT-4 to correct the errors in the transcript content. 
Last, the obtained raw dataset undergoes a thorough cleaning to ensure that the KOLs' privacy is not leaked.
To make our data suitable for most models, we translate all the data into English using GPT-4.
More details about the data can be found in the Appendix \ref{sec:appendix_dataset}.

These KOLs are experts in different fields, including skincare, fitness, and cooking. 
Each KOL's data includes 300 video transcripts and fans' comments. 
As shown in Table~\ref{tab:dataset}, compared to other personal and role-playing datasets, our data comes from the real world, and the video transcripts are narrated in the first-person perspective, making them closer to everyday life. 
Additionally, since the KOLs' opinions and knowledge are densely distributed in each video, playing the role of KOLs requires adhering to the long context formed by all the videos, which poses a challenge.



\subsection{RPLA Construction} 
\label{sec:methodology}




\paragraph{Data Preparation and Model Training} 
To develop RPLAs for KOLs, we first construct \textbf{meta-opinions} and \textbf{dialogues} for each KOL, sourced from the collected raw data. 
A rich knowledge base is a critical feature that distinguishes KOLs from character RPLAs. 
As shown in Figure \ref{tab:main_result}, to construct knowledge-intensive dialogue training data, we utilize GPT-4 to extract ten groups of \textbf{meta-opinions} \(\{v_{n1}, v_{n2}, ..., v_{n10}\}\) from each video transcript \(v_n\). 
Afterwards, for each group of meta-opinion, we then apply GPT-4 to simulate \textbf{dialogue} pairs between fans and KOLs \(\{\{c_{n11}, c_{n12}\}, \{c_{n21}, c_{n22}\}, ...\}\). 
Specifically, since the video transcripts are first-person perspectives, we require the model to simulate the KOL by responding as closely as possible to the original text. 
This approach ensures the KOL's tone is preserved while incorporating knowledge into the constructed data.
In addition, to mitigate the model's tendency to rely on parameter knowledge when answering questions, we use GPT-4 to determine whether the model's simulated KOL responses are consistent with GPT-4's direct responses to the same questions. 
This helps filter out those \textbf{counter-intuitive opinions}. 
We then concatenate these counter-intuitive data into the training data, enhancing the model's reliance on external knowledge.



In summary, our constructed training data includes two parts: knowledge-based dialogue data \(\{\{c_{n11}, c_{n12}\}, \{c_{n21}, c_{n22}\}, \ldots\}\) and counter-intuitive follow-up data \(\{v_{n1} + c_{n11}, c_{n12}\}, \{v_{n1} + c_{n21}, c_{n22}\}, \ldots\}\). 
We perform supervised fine-tuning (SFT) on qwen2-7b and qwen2-14b, with separate fine-tuning for each role to eliminate character hallucination problems caused by knowledge interference between characters.

\paragraph{Knowledge Retrieval}
During inference, to ensure accurate retrieval from extensive KOL knowledge, we employ retrieval-augmented generation.
We employ embedding-based retrieval, using OpenAI’s text-embedding-ada-002 as the retrieval model.
We segment each KOL's raw data into chunks of 500 tokens. 
The chunk with the highest relevance is matched for each query to assist the model's reasoning.

\subsection{Evaluation}
\label{sec:evaluation}
Unlike characters' RPLA, KOLs' domain knowledge is very dense. To comprehensively evaluate KOL RPLA's performance, we introduce two types of dimensions with distinct methodologies: basic performance evaluation and fan-centric evaluation.

\paragraph{Basic Performance Evaluation}
Similar to the evaluation of character RPLA\cite{tu2024charactereval}, we assess the basic performance of KOLs from two dimensions: tone characteristics capability and professional knowledge capability.
Tone characteristics aim to test whether the model can exhibit the speaking style of KOLs in communication, faithfully replicating each KOL's unique expression habits. 
Professional knowledge aims to test whether the model can generate knowledge consistent with KOLs, including unique opinions. 
To quantify the results, we adopt a multiple-choice evaluation method. 
Based on each video transcript, we ask GPT-4 to simulate user questions and KOL responses and design incorrect options to form multiple-choice questions. 
Specifically, we require each question to involve a certain reasoning process, and the incorrect options must be reasonable but not align with the KOL's speaking style or domain opinions. 
For example, a KOL in the health field believes that skipping breakfast is an effective way to stay healthy, which is not in line with mainstream opinions. 
We construct 500 questions for each KOL, covering all video transcripts.

\paragraph{Fan-Centric Evaluation}
\newcolumntype{a}{>{\columncolor{BlueGreen!10}}c}
\newcolumntype{b}{>{\columncolor{brown!10}}r}
\newcolumntype{d}{>{\columncolor{blue!10}}r}
\newcolumntype{q}{>{\columncolor{Green!10}}r}
\setlength\tabcolsep{2.8pt}
\begin{table*}[t]
\centering  
\footnotesize
\begin{tabular}{lbbqqqqdddd}
\toprule

\multicolumn{1}{c}{\multirow{4}[1]{*}{\textbf{Model}}} &  
\multicolumn{2}{c}{\textbf{Basic}} &
\multicolumn{8}{c}{\textbf{Fan-Centric}} \\

\cmidrule(lr){2-3}
\cmidrule(lr){4-11}

& \multicolumn{1}{c}{\multirow{3}{*}{\textbf{Know}}} &
\multicolumn{1}{c}{\multirow{3}{*}{\textbf{Tone}}} &
\multicolumn{4}{c}{\cellcolor{white}\textbf{New Fans}} &
\multicolumn{4}{c}{\cellcolor{white}\textbf{Old Fans}} \\

\cmidrule(lr){4-7}
\cmidrule(lr){8-11}

&\multicolumn{1}{c}{} & \multicolumn{1}{c}{} & \multicolumn{1}{c}{\cellcolor{white}\textbf{CC}} & \multicolumn{1}{c}\textbf{IA} & \multicolumn{1}{c}\textbf{EA} & \multicolumn{1}{c}\textbf{ALL} & \multicolumn{1}{c}\textbf{FR} & \multicolumn{1}{c}\textbf{CR} & \multicolumn{1}{c}\textbf{CA} & \multicolumn{1}{c}\textbf{ALL} \\

\midrule

\rowcolor[gray]{0.95} \multicolumn{11}{c}{\textit{Profile w/o RAG}} \\

\midrule
\addlinespace[0.05cm]

\textbf{GPT-3.5-Turbo} & 30.01 & 56.95 & 1.56 & 1.80 & 1.58 & 4.94 & 1.88 & 1.82 & 1.94 & 5.64  \\
\textbf{GPT-4o} & 29.83 & 60.41 & 1.70 & 2.22 & 1.81 & 5.73 & 2.26 & 2.14 & 2.10 & 6.50  \\
\textbf{Claude-3.5 Sonnect} & 30.11 & 61.53 & 1.69 & 2.23 & 1.83 & 5.75 & 2.30 & 2.15 & 2.10 & 6.55  \\
\textbf{Genmini-1.5 Pro} & 31.36 & 59.41 & 1.60 & 2.14 & 1.85 & 5.59 & 2.16 & 2.20 & 2.02 & 6.38  \\
\textbf{Qwen2-14B-Chat} & 30.57 & 49.88 & 1.50 & 1.42 & 1.55 & 4.47 & 1.71 & 1.57 & 1.88 & 5.16 \\
\textbf{Qwen2-72B-Chat} & 30.85 & 54.14 & 1.58 & 1.78 & 1.59 & 4.95 & 1.88 & 1.86 & 1.95 & 5.69 \\

\midrule

\rowcolor[gray]{0.95} \multicolumn{11}{c}{\textit{Profile w/ RAG}} \\

\textbf{GPT-3.5-Turbo} & 58.13 & 69.27 & 1.69 & 1.82 & 1.56 & 5.07 & 1.90 & 2.10 & 2.19 & 6.19  \\
\textbf{GPT-4o} & 70.80 & 88.20 & 1.76 & \textbf{2.35} & 1.88 & 5.99 & 2.36 & 2.46 & 2.32 & 7.14 \\
\textbf{Claude-3.5 Sonnect} & 70.92 & 82.91 & 1.77 & 2.20 & 1.86 & 5.83 & \textbf{2.39} & 2.27 & 2.18 & 6.84 \\
\textbf{Genmini-1.5 Pro} & 68.50 & 85.12 & 1.70 & 2.18 & 1.90 & 5.78 & 2.20 & 2.41 & 2.25 & 6.86 \\
\textbf{Qwen2-14B-Chat} & 55.25 & 68.41 & 1.59 & 1.52 & 1.61 & 4.72 & 1.79 & 1.90 & 2.09 & 5.78 \\
\textbf{Qwen2-72B-Chat} & 60.45 & 71.48 & 1.72 & 1.81 & 1.63 & 5.16 & 1.88 & 2.15 & 2.24 & 6.27 \\

\cdashlinelr{1-11}

\rowcolor[gray]{0.95} \multicolumn{11}{c}{\textit{sft w/o RAG}} \\

\textbf{Qwen2-7B-Chat} & 50.15 & 71.40 & 1.69 & 2.26 & 1.85 & 5.80 & 2.29 & 2.31 & 2.32 & 6.92 \\
\textbf{Qwen2-14B-Chat} & 59.89 & 78.93 & 1.74 & 2.30 & 1.90 & 5.94 & 2.34 & 2.38 & 2.40 & 7.12 \\

\cdashlinelr{1-11}

\rowcolor[gray]{0.95} \multicolumn{11}{c}{\textit{\method}} \\

\textbf{Qwen2-7B-Chat} & 68.90 & 86.40 & 1.74 & 2.32 & 1.88 & 5.94 & 2.31 & 2.49 & 2.35 & 7.15 \\
\textbf{Qwen2-14B-Chat} & \textbf{73.91} & \textbf{89.15} & \textbf{1.80} & 2.34 & \textbf{1.92} & \textbf{6.02} & 2.35 & \textbf{2.52} & \textbf{2.42} & \textbf{7.29} \\

\bottomrule
\end{tabular}
\caption{The results of different models and methods on two evaluation tasks. Basic Performance Evaluation uses accuracy as the metric, while Fan-Centric Evaluation uses the scores given by GPT-4 as the metric. \textbf{ALL} represents the sum of three metrics for one type of fan.}
\label{tab:main_result}
\vspace{-0.2cm}
\end{table*}

Fan-centric is a significant feature of KOLs. KOLs often adjust their content and interaction methods based on the needs and feedback of different fans. 
From this perspective, we model two types of users who might interact with the KOL RPLA: new users getting to know KOLs (new fans) and loyal fans of KOLs (old fans).
To better model the latter, we use fan comments from the raw data as source data, providing them to GPT-4 to construct profiles of this fan group. 
These profiles include their age range, interests, lifestyle, career tendencies, consumption habits, language style, etc. 
We then have the user RPLAs engage in five rounds of interaction with the KOL's RPLA based on specific video content and use GPT-4 to evaluate the interaction. 
For new fans, we provide the following three dimensions for evaluation:

\begin{itemize}
    \item \textbf{Content Comprehension (CC)}: Evaluate whether the content provided by KOLs ensures accuracy while also making it easier for new fans to understand and grasp.
    \item \textbf{Interaction Attractiveness (IA)}: Evaluate the RPLA's effectiveness in interacting with new fans, including its ability to respond reasonably, interact friendly, and make new fans feel valued and cared for.
    \item \textbf{Engagement Appeal (EA)}: Evaluate whether the RPLA can sustain new fans' interest and encourage further engagement during interactions. This involves content enjoyment, interaction guidance, and attracting and converting new users into long-term fans.
\end{itemize}

For old fans, we provide the following three dimensions for evaluation:
\begin{itemize}
    \item \textbf{Fan Resonance (FR)}: Evaluate if the RPLA's interaction with old fans elicits emotional resonance. Observe if it understands and responds to their emotional needs and concerns and if this interaction enhances fan loyalty and connection.
    \item \textbf{Content Relevance (CR)}: Evaluate if the RPLA content aligns with old fans' interests and needs, ensuring it meets their expectations and maintains their attention and support for the KOL.
    \item \textbf{Character Authenticity (CA)}: Evaluate if the RPLA  authentically replicates the KOL's personality and style during interactions with old fans. Ensure the RPLA's behavior, language, and attitude are consistent with the KOL's usual performance, providing old fans with familiarity and continuity.
\end{itemize}

To rate these KOL RPLAs,  we employ GPT-4 to assign scores of 1 (Poor), 2 (Average), and 3 (Excellent), providing detailed descriptions for different scores across each dimension. Specific prompts are detailed in Appendix \ref{sec:appendix_prompt_eval}.

\begin{table*}[t]
  \centering
  \small
    \begin{tabularx}{\linewidth}{X}

    \toprule
    
    \textbf{Question}: Could you share any good methods for protecting the skin barrier?  \\
    
    \toprule
    
    \textbf{GPT-3.5} \hspace{20pt} Profile w/ RAG \\

    \midrule

    Hello dear! Protecting the skin barrier is very important, especially when facing external environmental stress and pollution. Here are some methods I recommend:
        1. Gentle cleansing: Choose mild cleansing products to avoid over-cleansing and damaging the skin barrier.
        2. Use moisturizers: \textcolor{teal}{Select skincare products containing moisturizing ingredients like hyaluronic acid and glycerin to maintain skin hydration.}
        3. Sunscreen: Use sunscreen with SPF daily to prevent UV damage.
        4. Avoid irritants: Try to avoid using skincare products containing alcohol and fragrances, as these can irritate the skin. \\

    \midrule

    \textbf{GPT-4} \hspace{20pt} Profile w/ RAG \\

    \midrule

    Hello dear! Regarding protecting the skin barrier, I have a few suggestions that I hope will be helpful to you: \\
1. Gentle cleansing: Avoid using strongly irritating cleansers, and choose mild face wash products that are free from soap bases, alcohol, and artificial fragrances. \\
2. Moisturize and lock in water: Moisturizing is key. Choose moisturizing products suitable for your skin type, \textcolor{teal}{especially those containing ceramides, hyaluronic acid, and glycerin, which help maintain the integrity of the skin barrier.} \\
3. Proper use of masks: Although masks can provide instant hydration, frequent use or improper selection can burden the skin. Choose masks suitable for sensitive skin, and limit their use to 2-3 times a week. \\
4. Repair products: Consider specialized repair products if your skin barrier is already damaged. \textcolor{brown}{Here, I highly recommend the URIAGE brand's rich cream. It can quickly alleviate redness and discomfort and enhance the skin's defense capabilities.} \\
5. Healthy lifestyle: Maintain a healthy diet rich in vitamins and antioxidants, such as fresh fruits and vegetables. Adequate sleep and moderate exercise also help keep the skin healthy and vibrant. \\
I hope these methods are helpful to you! If you have more questions, feel free to contact me anytime. \\ 

    \midrule

    \textbf{\method} \\

    \midrule

    Hello darling! I am happy to receive your question, and I have many secrets to share about protecting the skin barrier! \textcolor{teal}{First, we need to understand that the skin barrier includes both a physical barrier a defense barrier, and a microbial barrier.} These three components work together to protect our skin's health. Therefore, building a comprehensive skincare routine is very important! Now, let's talk about how to protect our skin barrier: \\
1. \textbf{Gentle cleansing}: Choose gentle cleansers that do not contain irritating ingredients and avoid over-cleansing. I know many people now adhere to the routine of using \textcolor{red}{morning C night A}, \textcolor{brown}{but this may not necessarily be effective for your skin barrier. I prefer a simplified skincare routine, using fewer unnecessary ingredients, and washing my face with water is best.} \\
2. \textbf{Proper moisturizing}: Choose moisturizing products suitable for your skin type to ensure skin hydration. \textcolor{brown}{The URIAGE rich cream is a good choice. It contains 300DAIRU rapid soothing technology, which can quickly reduce redness and relieve discomfort. Additionally, it contains various plant oils, making it very suitable for daily maintenance of dry and sensitive skin.} \\
3. \textbf{Comprehensive repair}: Apart from the physical barrier, we must also repair the defense and microbial barriers. \textcolor{brown}{URIAGE products can help us achieve this. Their rich cream helps the skin resist external stimuli and enhances self-regulation capabilities.} \\
4. \textbf{Ingredient selection}: Be cautious of "pseudo-repair" products that seem effective but may cause dependency with long-term use. \textcolor{brown}{Brands like URIAGE provide genuinely effective repair ingredients that can truly enhance skin tolerance.} \\
5. \textbf{Maintaining healthy habits}: Good sleep and dietary habits are important for skin health. Remember to drink plenty of water, eat more fruits and vegetables, and get enough sleep. \textcolor{red}{Your skin condition will take off.} \\
In summary, you should insist on gentle care and comprehensive repair, thus making your skin barrier healthier and more resilient. \textcolor{brown}{Take care, and beauty will find you!} If you have more questions, feel free to leave a message, and I'll do my best to help you.
I look forward to hearing about your skincare progress~ \\
    
    \bottomrule
    
    \end{tabularx}
\caption{Case study results of different methods. \textcolor{teal}{Teal} represents the opinions of the LLMs, \textcolor{brown}{Brown} represents the personal opinions and catchphrases of the KOL, and \textcolor{Red}{Red} represents some emerging internet slang.}
\label{table:casestudy}
\end{table*}

\section{Experiment}

\subsection{Baseline}
We fine-tune the model by constructing knowledge-intensive training data and enhancing it through RAG. To validate the effectiveness of this methodology and its different components, we adopt baselines in three settings:
\begin{itemize}
    \item \textit{In-contexting (Profile w/o RAG)}: Only provide the model with the KOL's profile \( p \), allowing it to play the role of the corresponding KOL directly.
    \item \textit{In-contexting (Profile w/ RAG)}: Provide the model with the KOL's profile and RAG content, allowing it to respond to users while referring to RAG content.
    \item \textit{Supervised Fine-Tuning (sft)}: The model is trained using \method but without providing RAG content.
\end{itemize}

For methods without training, we conduct evaluations on GPT-3.5\cite{openai2022chatgpt}, GPT-4\cite{openai2023gpt4}, Claude-3.5\cite{anthropic2024claude3}, Gemini1.5-Pro\cite{geminiteam2024gemini}, Qwen2-14B-Chat\cite{qwen}, and Qwen2-72B-Chat. For the SFT method, we conduct evaluations on Qwen2-7B-Chat and Qwen2-14B-Chat
\footnote{The versions in this paper are \texttt{gpt-3.5-turbo-0125}, \texttt{gpt-4o-2024-08-06}, \texttt{gemini-1.5-pro} and \texttt{claude-3.5-sonnet-20240620} respectively.}.

\subsection{Main Result}

As shown in Table~\ref{tab:main_result}, from the evaluation of basic performance, we can observe the following: First, when only providing a profile, the model's results in knowledge Q\&A are close to random (25\%). On the one hand, this indicates that LLMs have difficulty answering domain-specific questions based on knowledge imbibed in their parameters. On the other hand, unlike other RPLA works, this means that opinions from KOL do not face the issue of data leakage.
Second, the results improve after integrating RAG, regardless of whether the model has been trained. 
This demonstrates the necessity of RAG. 
The improvement is less pronounced after training, which indicates that the model has already learned a substantial amount of KOL's knowledge during the training period.
Last, \method based on qwen-14b outperforms GPT-4 w/ RAG in most metrics, demonstrating the effectiveness of our approach.
From the Fan-Centric evaluation results, it can be seen that GPT-4 performs similarly to \method on the metrics CC, IE, FR, and EA, which can be attributed to its emotional solid expression capabilities. However, it performs poorly on metrics such as CR and CA, indicating its ineffectiveness in critical knowledge-based expression, which is important for KOL. 
Sec \ref{sec:case} reveals this because GPT-4 tends to provide its own opinions in responses rather than that of the KOL.


\begin{table}[t]
    \centering
    \small
    \begin{tabular}{lccc}
        \toprule
        \textbf{LLMs} & \textbf{training} & \textbf{Knowledge} & \textbf{Tone} \\
        \midrule
        GPT-3.5 & \ding{55} & 65.93(58.13) & 82.24(69.27) \\
        GPT-4 & \ding{55} & 88.24(70.80) & 90.34(88.20) \\
        qwen2-14b & \ding{55} & 63.23(55.25) & 80.51(68.41) \\
        qwen2-14b & \ding{51} & \textbf{89.13(73.91)} & \textbf{92.14(89.15)} \\
        \bottomrule
    \end{tabular}
    \caption{Analysis of the impact of Reference content and Retrieval content on the results. Inside and outside the () are the retrieval and reference content results, respectively.}
    \label{tab:retrieval}
\end{table}

\subsection{Reference Content v.s. Retrieval Content}
To analyze the impact of retrieved content on the model, we provide reference video content for each question. 
As shown in Table~\ref{tab:retrieval}, both tone and knowledge tasks improve when given reference results, particularly noticeable in the knowledge tasks. This indicates that there is still room for improvement in the retrieval methods.
For the tone tasks, answers often directly cite the original words from the video, while knowledge tasks require some level of reasoning. 
Therefore, the knowledge tasks are not always correctly answered, even with reference content. 
Moreover, in the Language tasks, models through training consistently outperform those using in-context learning. 
This suggests that models understand the roles more deeply after training, enabling them to better infer the characters' language patterns without reference to content.

\begin{table}[t]
    \centering
    \small
    \begin{tabular}{llcc}
        \toprule
        \textbf{LLMs} & \textbf{Method} & \textbf{Knowledge} & \textbf{Tone} \\
        \midrule
        Claude-3.5 & Profile w/ RAG & 70.92 & 82.91 \\
        Claude-3.5 & long-context & \textbf{71.09} & \textbf{90.45} \\
        Genmini-1.5 Pro & Profile w/ RAG & 68.50 & 85.12 \\
        Genmini-1.5 Pro & long-context & 70.25 & 86.07 \\
        \bottomrule
    \end{tabular}
    \caption{The results of using long-context models for KOL's RPLA, where the long-context method incorporates all video content as the model's context.}
    \label{tab:longcontext}
\end{table}

\subsection{Long-context Model with All Video Content }
Directly using long-context models for role-playing is an interesting topic. 
In the knowledge-intensive KOL RPLA, incorporating all video content as context helps evaluate the model's ability to portray the overall character and thoroughly assesses the model's ability to retrieve knowledge from the context. 
This is because our questions originate from various segments within the context and require a certain level of reasoning. 
As shown in Table~\ref{tab:longcontext}, we evaluate two representative long-context models, Claude-3.5-sonnet and Gemini-1.5-pro. 
Although their performance is not as strong as \method, long-context models still show potential for role-playing, especially in free dialogue scenarios.
Their performance in knowledge tasks is worse than in tone tasks, possibly because effective free dialogue relies more on an overall understanding of the character and can tolerate some context loss.

\subsection{Case Study}
\label{sec:case}
As shown in Table~\ref{table:casestudy}, we analyze the results of profile w/o RAG, profile w/ RAG, and \method on the fan-centric evaluation. 
It can be observed that GPT-3.5's responses are relatively brief and often derived from its parameter knowledge. 
In contrast, GPT-4 provides more detailed responses and follows our instructions by incorporating retrieved knowledge into its answers.
However, the tone of GPT-4's responses does not resemble that of the KOL, and it still relies on its knowledge sometimes. 
The model trained using \method more extensively includes the KOL's knowledge while adhering to the KOL's linguistic habits. 
For example, the phrase "Take care, and beauty will find you!" is a common saying among KOLs, and its appearance in responses enhances the authenticity of the role-playing. 
Furthermore, the responses from the \method include some emerging internet slang, indicating that role-playing as a KOL requires considering the complexity and variability of the real world. The model needs to understand these terms and incorporate them into its responses.

\setlength\tabcolsep{3pt}
\begin{table}[t]
\small
  \centering
    \begin{tabular}{lccc}
    \toprule
    \textbf{LLMs} & \textbf{Pearson's $r$} & \textbf{Spearman's $\rho$} & \textbf{Kendall's $\tau$}  \\
    \midrule
    \rowcolor[rgb]{ .949,  .953,  .961} \multicolumn{4}{c}{\textit{New fans}} \\
     GPT-4  & 83.52 & 85.91 & 76.42\\  
    qwen2-14b  & 84.00 & 85.91 & 77.52\\
     
    \midrule

    \rowcolor[rgb]{ .949,  .953,  .961} \multicolumn{4}{c}
    {\textit{Old fans}} \\
     GPT-4  & 79.55 & 79.92 & 75.12\\  
    qwen2-14b  & 79.20 & 80.31 & 75.59\\
         
    \bottomrule
    \end{tabular}
  \caption{Comparison between GPT-4 evaluation and human evaluation on the fan-centric task.}
  \label{tab:human_eval}
\end{table}
\subsection{Human Evaluation}
We also conduct a human evaluation for the Fan-Centric Evaluation to validate GPT-4's performance on this task. 
We compare the predictions from the LLMs with human judgments. 
To evaluate new and old fans, we sample 100 sets of dialogue data from 5 KOLs and enlist 6 crowd-sourced workers to annotate the data. 
Three annotators who have not watched KOL videos evaluate the dialogues of new fans, while the other three, who have watched some KOL videos, evaluate the dialogues of old fans. 
All annotators follow the same criteria as GPT-4 and are compensated at the local minimum wage. 
We report the Pearson’s $r$~\citep{pearson1920notes}, Spearman's $\rho$~\citep{spearman1961proof} and Kendall's $\tau$~\citep{kendall1938new} correlations between human annotations and GPT-4.
As shown in Table~\ref{tab:human_eval}, GPT-4's evaluations closely match the human evaluations. The simulation of new fans is better compared to old fans.
We believe this is because the annotators are not experts in the KOL's domain and cannot fully simulate such individuals.
\section{Conclusion}

In this paper, we study the RPLA of KOLs based on language models. 
KOLs are characterized by real-world grounding, knowledge intensity, and avoidance of data leakage, making it a more complex role-playing task.
We propose a complete framework \method that spans data collection and model training to evaluation. 
For data collection, we constructed the first KOL dataset using many video transcripts. 
For model training, we implemented targeted optimizations to create training data.
For evaluation, we innovatively proposed a fan-centric evaluation scheme to assess the KOL RPLA better. Experimental results demonstrate the advantages of our approach. 

\section*{Limitations}
\label{sec:limitation}
In this paper, we primarily focus on Key Opinion Leaders(KOLs) role-playing. 
KOLs typically present themselves as experts in a specific field, which differs to some extent from genuine human data, as they often maintain certain personas online. 
Furthermore, our data mainly comes from the Chinese internet and has distinct cultural and regional characteristics, necessitating further data expansion to eliminate these biases.

Another limitation of our work comes from potential biases in the evaluation process. 
Although we chose video transcript data, KOL-related content may still appear in some models' training data. Additionally, the evaluation metrics used in this paper rely on evaluator LLMs, which may compromise the accuracy of results due to inherent errors in these models, leading to biased evaluations of profile consistency.
Furthermore, our Retrieval-Augmented Generation (RAG) system uses the most basic retrieval scheme, which has the potential for improvement to maximize the ability of \method to role-play KOLs.

\section*{Ethics Statement}
\label{sec:Ethics}

\paragraph{Use of Human Annotations}
Our institution recruits annotators to implement the annotations of motivation recognition dataset construction. 
We ensure the privacy rights of the annotators are respected during the annotation process.
The annotators receive compensation exceeding the local minimum wage and have consented to using motivation recognition data they process for research purposes. Appendix~\ref{sec:appendix_human} provides further details on the annotations.

\paragraph{Risks}
The \method dataset is sourced from KOL video transcripts. However, we cannot guarantee they are devoid of socially harmful or toxic language. Additionally, we have implemented some automated methods to prevent the leakage of KOLs' personal information in the data, but some lapses may still occur. Moreover, our evaluation metrics are based on related books on KOL research, but they may vary due to timeliness and the diverse backgrounds of individuals.


\bibliography{anthology,custom}

\begin{thebibliography}{36}
\expandafter\ifx\csname natexlab\endcsname\relax\def\natexlab#1{#1}\fi

\bibitem[{Ahn et~al.(2023)Ahn, Song, Yun, and Kim}]{ahn2023mpchat}
Jaewoo Ahn, Yeda Song, Sangdoo Yun, and Gunhee Kim. 2023.
\newblock \href {http://arxiv.org/abs/2305.17388} {Mpchat: Towards multimodal persona-grounded conversation}.

\bibitem[{Amalia et~al.(2023)}]{amalia2023effect}
Amanda~Risky Amalia et~al. 2023.
\newblock The effect of key opinion leader, instagram, and digital campaign through brand image on brand awareness in evoria products.
\newblock \emph{JMKSP (Jurnal Manajemen, Kepemimpinan, dan Supervisi Pendidikan)}, 8(2):1094--1111.

\bibitem[{Anthropic(2023)}]{Anthropic2023Claude2}
Anthropic. 2023.
\newblock \href {https://www-files.anthropic.com/production/images/Model-Card-Claude-2.pdf} {Model card and evaluations for claude models}.

\bibitem[{Anthropic(2024)}]{anthropic2024claude3}
Anthropic. 2024.
\newblock \href {https://www-cdn.anthropic.com/de8ba9b01c9ab7cbabf5c33b80b7bbc618857627/Model_Card_Claude_3.pdf} {The claude 3 model family: Opus, sonnet, haiku}.

\bibitem[{Bai et~al.(2023)Bai, Bai, Chu, Cui, Dang, Deng, Fan, Ge, Han, and Huang}]{qwen}
Jinze Bai, Shuai Bai, Yunfei Chu, Zeyu Cui, Kai Dang, Xiaodong Deng, Yang Fan, Wenbin Ge, Yu~Han, and Fei Huang. 2023.
\newblock Qwen technical report.
\newblock \emph{arXiv preprint arXiv:2309.16609}.

\bibitem[{Bamakan et~al.(2019)Bamakan, Nurgaliev, and Qu}]{bamakan2019opinion}
Seyed Mojtaba~Hosseini Bamakan, Ildar Nurgaliev, and Qiang Qu. 2019.
\newblock Opinion leader detection: A methodological review.
\newblock \emph{Expert Systems with Applications}, 115:200--222.

\bibitem[{Chen et~al.(2024)Chen, Wang, Xu, Yuan, Zhang, Shi, Xie, Li, Yang, Zhu et~al.}]{chen2024persona}
Jiangjie Chen, Xintao Wang, Rui Xu, Siyu Yuan, Yikai Zhang, Wei Shi, Jian Xie, Shuang Li, Ruihan Yang, Tinghui Zhu, et~al. 2024.
\newblock From persona to personalization: A survey on role-playing language agents.
\newblock \emph{arXiv preprint arXiv:2404.18231}.

\bibitem[{Desai et~al.(2021)Desai, Kaul, Aysola, and Johnson}]{desai2021redcapswebcuratedimagetextdata}
Karan Desai, Gaurav Kaul, Zubin Aysola, and Justin Johnson. 2021.
\newblock \href {http://arxiv.org/abs/2111.11431} {Redcaps: web-curated image-text data created by the people, for the people}.

\bibitem[{Gao et~al.(2023)Gao, Lian, Zhou, Fu, and Wang}]{gao2023livechat}
Jingsheng Gao, Yixin Lian, Ziyi Zhou, Yuzhuo Fu, and Baoyuan Wang. 2023.
\newblock Livechat: A large-scale personalized dialogue dataset automatically constructed from live streaming.
\newblock \emph{arXiv preprint arXiv:2306.08401}.

\bibitem[{Gao et~al.(2024)Gao, Xiong, Gao, Jia, Pan, Bi, Dai, Sun, Guo, Wang, and Wang}]{gao2024retrievalaugmented}
Yunfan Gao, Yun Xiong, Xinyu Gao, Kangxiang Jia, Jinliu Pan, Yuxi Bi, Yi~Dai, Jiawei Sun, Qianyu Guo, Meng Wang, and Haofen Wang. 2024.
\newblock \href {http://arxiv.org/abs/2312.10997} {Retrieval-augmented generation for large language models: A survey}.

\bibitem[{Jang et~al.(2022)Jang, Lim, Hur, Oh, Son, Lee, Shin, Kim, and Lim}]{jang2022customized}
Yoonna Jang, Jungwoo Lim, Yuna Hur, Dongsuk Oh, Suhyune Son, Yeonsoo Lee, Donghoon Shin, Seungryong Kim, and Heuiseok Lim. 2022.
\newblock \href {http://arxiv.org/abs/2112.08619} {Call for customized conversation: Customized conversation grounding persona and knowledge}.

\bibitem[{Kendall(1938)}]{kendall1938new}
Maurice~G Kendall. 1938.
\newblock A new measure of rank correlation.
\newblock \emph{Biometrika}, 30(1/2):81--93.

\bibitem[{Li et~al.(2023)Li, Leng, Yan, Shen, Wang, MI, Fei, Feng, Yan, Wang et~al.}]{li2023chatharuhi}
Cheng Li, Ziang Leng, Chenxi Yan, Junyi Shen, Hao Wang, Weishi MI, Yaying Fei, Xiaoyang Feng, Song Yan, HaoSheng Wang, et~al. 2023.
\newblock Chatharuhi: Reviving anime character in reality via large language model.
\newblock \emph{arXiv preprint arXiv:2308.09597}.

\bibitem[{Lin(2004)}]{lin2004rouge}
Chin-Yew Lin. 2004.
\newblock Rouge: A package for automatic evaluation of summaries.
\newblock In \emph{Text summarization branches out}, pages 74--81.

\bibitem[{OpenAI(2022)}]{openai2022chatgpt}
OpenAI. 2022.
\newblock \href {https://openai.com/blog/chatgpt} {Chatgpt}.

\bibitem[{OpenAI(2023)}]{openai2023gpt4}
OpenAI. 2023.
\newblock \href {http://arxiv.org/abs/2303.08774} {Gpt-4 technical report}.

\bibitem[{Palmer and Supuran(2015)}]{palmer2015key}
Claudia Palmer and Claudiu~T Supuran. 2015.
\newblock A key opinion leader interview: insight into the research and career of prof. claudiu t supuran.
\newblock \emph{Expert Opinion on Therapeutic Patents}, 25(4):501--505.

\bibitem[{Park et~al.(2023)Park, O'Brien, Cai, Morris, Liang, and Bernstein}]{park2023generative}
Joon~Sung Park, Joseph~C. O'Brien, Carrie~J. Cai, Meredith~Ringel Morris, Percy Liang, and Michael~S. Bernstein. 2023.
\newblock \href {http://arxiv.org/abs/2304.03442} {Generative agents: Interactive simulacra of human behavior}.

\bibitem[{Pearson(1920)}]{pearson1920notes}
Karl Pearson. 1920.
\newblock Notes on the history of correlation.
\newblock \emph{Biometrika}, 13(1):25--45.

\bibitem[{Radford et~al.(2022)Radford, Kim, Xu, Brockman, McLeavey, and Sutskever}]{radford2022robustspeechrecognitionlargescale}
Alec Radford, Jong~Wook Kim, Tao Xu, Greg Brockman, Christine McLeavey, and Ilya Sutskever. 2022.
\newblock \href {http://arxiv.org/abs/2212.04356} {Robust speech recognition via large-scale weak supervision}.

\bibitem[{Ran et~al.(2024)Ran, Wang, Xu, Yuan, Liang, Xiao, and Yang}]{ran2024capturing}
Yiting Ran, Xintao Wang, Rui Xu, Xinfeng Yuan, Jiaqing Liang, Yanghua Xiao, and Deqing Yang. 2024.
\newblock Capturing minds, not just words: Enhancing role-playing language models with personality-indicative data.
\newblock \emph{arXiv preprint arXiv:2406.18921}.

\bibitem[{Salemi et~al.(2024)Salemi, Mysore, Bendersky, and Zamani}]{salemi2024lamp}
Alireza Salemi, Sheshera Mysore, Michael Bendersky, and Hamed Zamani. 2024.
\newblock \href {http://arxiv.org/abs/2304.11406} {Lamp: When large language models meet personalization}.

\bibitem[{Sellam et~al.(2020)Sellam, Das, and Parikh}]{sellam2020bleurt}
Thibault Sellam, Dipanjan Das, and Ankur~P Parikh. 2020.
\newblock Bleurt: Learning robust metrics for text generation.
\newblock \emph{arXiv preprint arXiv:2004.04696}.

\bibitem[{Shao et~al.(2023)Shao, Li, Dai, and Qiu}]{shao2023character}
Yunfan Shao, Linyang Li, Junqi Dai, and Xipeng Qiu. 2023.
\newblock \href {https://aclanthology.org/2023.emnlp-main.814} {Character-{LLM}: A trainable agent for role-playing}.
\newblock In \emph{Proceedings of the 2023 Conference on Empirical Methods in Natural Language Processing}, pages 13153--13187, Singapore. Association for Computational Linguistics.

\bibitem[{Spearman(1961)}]{spearman1961proof}
Charles Spearman. 1961.
\newblock The proof and measurement of association between two things.

\bibitem[{Team(2024)}]{geminiteam2024gemini}
Gemini Team. 2024.
\newblock \href {http://arxiv.org/abs/2312.11805} {Gemini: A family of highly capable multimodal models}.

\bibitem[{Touvron et~al.(2023)Touvron, Martin, Stone, Albert, Almahairi, Babaei, and Bashlykov}]{touvron2023llama}
Hugo Touvron, Louis Martin, Kevin Stone, Peter Albert, Amjad Almahairi, Yasmine Babaei, and Nikolay Bashlykov. 2023.
\newblock \href {http://arxiv.org/abs/2307.09288} {Llama 2: Open foundation and fine-tuned chat models}.

\bibitem[{Tu et~al.(2024)Tu, Fan, Tian, and Yan}]{tu2024charactereval}
Quan Tu, Shilong Fan, Zihang Tian, and Rui Yan. 2024.
\newblock Charactereval: A chinese benchmark for role-playing conversational agent evaluation.
\newblock \emph{arXiv preprint arXiv:2401.01275}.

\bibitem[{Urbanek et~al.(2019)Urbanek, Fan, Karamcheti, Jain, Humeau, Dinan, Rocktäschel, Kiela, Szlam, and Weston}]{urbanek2019learning}
Jack Urbanek, Angela Fan, Siddharth Karamcheti, Saachi Jain, Samuel Humeau, Emily Dinan, Tim Rocktäschel, Douwe Kiela, Arthur Szlam, and Jason Weston. 2019.
\newblock \href {http://arxiv.org/abs/1903.03094} {Learning to speak and act in a fantasy text adventure game}.

\bibitem[{Wang et~al.(2024)Wang, Xiao, tse Huang, Yuan, Xu, Guo, Tu, Fei, Leng, Wang, Chen, Li, and Xiao}]{wang2023does}
Xintao Wang, Yunze Xiao, Jen tse Huang, Siyu Yuan, Rui Xu, Haoran Guo, Quan Tu, Yaying Fei, Ziang Leng, Wei Wang, Jiangjie Chen, Cheng Li, and Yanghua Xiao. 2024.
\newblock \href {http://arxiv.org/abs/2310.17976} {Incharacter: Evaluating personality fidelity in role-playing agents through psychological interviews}.

\bibitem[{Wang et~al.(2023)Wang, Peng, Que, Liu, Zhou, Wu, Guo, Gan, Ni, Zhang et~al.}]{wang2023rolellm}
Zekun~Moore Wang, Zhongyuan Peng, Haoran Que, Jiaheng Liu, Wangchunshu Zhou, Yuhan Wu, Hongcheng Guo, Ruitong Gan, Zehao Ni, Man Zhang, et~al. 2023.
\newblock Rolellm: Benchmarking, eliciting, and enhancing role-playing abilities of large language models.
\newblock \emph{arXiv preprint arXiv:2310.00746}.

\bibitem[{Xu et~al.(2024)Xu, Wang, Chen, Yuan, Yuan, Liang, Chen, Dong, and Xiao}]{xu2024character}
Rui Xu, Xintao Wang, Jiangjie Chen, Siyu Yuan, Xinfeng Yuan, Jiaqing Liang, Zulong Chen, Xiaoqing Dong, and Yanghua Xiao. 2024.
\newblock Character is destiny: Can large language models simulate persona-driven decisions in role-playing?
\newblock \emph{arXiv preprint arXiv:2404.12138}.

\bibitem[{Yuan et~al.(2024)Yuan, Yuan, Cui, Lin, Wang, Xu, Chen, and Yang}]{yuan2024evaluating}
Xinfeng Yuan, Siyu Yuan, Yuhan Cui, Tianhe Lin, Xintao Wang, Rui Xu, Jiangjie Chen, and Deqing Yang. 2024.
\newblock Evaluating character understanding of large language models via character profiling from fictional works.
\newblock \emph{arXiv preprint arXiv:2404.12726}.

\bibitem[{Zhang et~al.(2018)Zhang, Dinan, Urbanek, Szlam, Kiela, and Weston}]{zhang2018personalizing}
Saizheng Zhang, Emily Dinan, Jack Urbanek, Arthur Szlam, Douwe Kiela, and Jason Weston. 2018.
\newblock \href {http://arxiv.org/abs/1801.07243} {Personalizing dialogue agents: I have a dog, do you have pets too?}

\bibitem[{Zhang et~al.(2019)Zhang, Kishore, Wu, Weinberger, and Artzi}]{zhang2019bertscore}
Tianyi Zhang, Varsha Kishore, Felix Wu, Kilian~Q Weinberger, and Yoav Artzi. 2019.
\newblock Bertscore: Evaluating text generation with bert.
\newblock \emph{arXiv preprint arXiv:1904.09675}.

\bibitem[{Zhao et~al.(2019)Zhao, Peyrard, Liu, Gao, Meyer, and Eger}]{zhao2019moverscore}
Wei Zhao, Maxime Peyrard, Fei Liu, Yang Gao, Christian~M Meyer, and Steffen Eger. 2019.
\newblock Moverscore: Text generation evaluating with contextualized embeddings and earth mover distance.
\newblock \emph{arXiv preprint arXiv:1909.02622}.

\end{thebibliography}
\clearpage
\appendix
\definecolor{Gray}{gray}{0.9}
\begin{table*}[ht]
    \centering
    \begin{tabular}{|p{0.3\linewidth}|p{0.3\linewidth}|p{0.3\linewidth}|}
        \hline
        \multicolumn{3}{|c|}{\textbf{The domain of 30 KOLs}} \\ \hline
        \rowcolor{Gray} Milk Tea Development & Psychological Counseling & Technology Insights \\ \hline
        Urban Exploration & Aviation & Car Reviews \\ \hline
        \rowcolor{Gray} Cross-border Law & Relationship Advice & Emotional Support \\ \hline
        Financial Management & Business English & lifestyle vlogging \\ \hline
        \rowcolor{Gray} Automotive Commentary & Growth and Learning & Esports \\ \hline
        Frontend Development & Fitness Coaching & Writing Guidance \\ \hline
        \rowcolor{Gray} Gaokao Arts Study & Efficient Cooking & Skincare \\ \hline
        Beauty and Aesthetic Medicine & Culinary Sharing & Midnight Food Exploration \\ \hline
        \rowcolor{Gray} Oily Acne Skin Care & Traditional Culinary Culture & Sailing \\ \hline
        Feminism & Nutrition and Health & Stand-up Comedy \\ \hline
    \end{tabular}
    \caption{Different Domains of 30 KOLs, each with their own specialized area, without any overlap.}
    \label{tab:kol_domain}
\end{table*}

\begin{table*}[t]
  \centering
  \small
    \begin{tabular}{l}
    \toprule
    \rowcolor[gray]{0.95}\multicolumn{1}{c}{\textbf{Prompt I}} \\
    \makecell[l]{
Your task is constructing ten sets of dialogue data based on a video transcript from a Key Opinion Leader (KOL). You need\\ to simulate conversations between fans, and the KOL centered around the KOL's personal opinions. \\
\color[rgb]{0,0.39,0}{\# Steps for Generation:}\\
Step 1: Read KOL's profile and the video transcript.\\
Step 2: Generate a "meta-opinion," which is the personal opinion the KOL expresses in the video transcript relating to the\\ KOL’s specific field.\\
Step 3: Simulate conversations between the KOL and their fans based on the generated "meta-opinion." The conversations\\ should focus on the "meta-opinion."\\
\color[rgb]{0,0.39,0}{\# Notes:}\\
1. The KOL's responses should closely mimic the transcript's original language or use language that matches the KOL’s style.\\
2. The "meta-opinion" should ideally be a relatively niche viewpoint.\\
3. Fans' questions should not involve politics or personal privacy.\\
\color[rgb]{0,0.39,0}{\# KOL's profile:}\\
\{profile\}\\
\color[rgb]{0,0.39,0}{\# Video transcript content:}\\
\{video\}\\
\color[rgb]{0,0.39,0}{\# Return format:}\\
The result should be returned as a JSON with ten dictionaries, including the generated ‘meta-opinion’ and the simulated\\ conversations between fans and KOL labeled as “fans” and “KOL”.}\\
    \bottomrule
    \end{tabular}
  \caption{Prompt templates for constructing initial training data.}
  \label{tab:template_for_data_construction}
\end{table*}


    

\begin{table}[t]
\centering  
\footnotesize
\begin{tabular}{lccccc}
\toprule
\multicolumn{1}{c}{\multirow{2}[1]{*}{\makecell{\textbf{Method}}}} &  
\multicolumn{4}{c}{\textbf{Evaluation Metric}} \\

\cmidrule(lr){2-6}
& \multicolumn{1}{c}\textbf{ROUGE} & \multicolumn{1}{c}\textbf{BERT.} & \multicolumn{1}{c}\textbf{BLRT.} & \multicolumn{1}{c}\textbf{Mover.}\\
\midrule
\addlinespace[0.05cm]
GPT-3.5 w/o RAG & 10.41	& 45.14	& 30.55	& 40.29 \\
GPT-4 w/o RAG & 18.03 & 50.22 & 41.14 & 43.86 \\
GPT-3.5 w/ RAG & 24.15 & 56.99 & 42.51 & 50.35 \\
GPT-4 w/ RAG & 30.82 & 58.45 & 49.13 & 56.72 \\
Qwen2-14B w/o RAG & 37.89 & 58.00 & 53.40 & 59.39 \\
\method & \textbf{39.08} & \textbf{61.51} & \textbf{55.25} & \textbf{61.99} \\
\bottomrule
\end{tabular}
\caption{Evaluation of KOL RPLA results using traditional automated evaluation metrics.The results for the GPT series models are obtained using in-context learning, while the results for Qwen2-14B are obtained through supervised fine-tuning.}
\label{tab:tradition}
\vspace{-0.2cm}
\end{table}

\begin{table*}[t]
  \centering
  \small
    \begin{tabular}{l}
    \toprule
    \rowcolor[gray]{0.95}\multicolumn{1}{c}{\textbf{Prompt II}} \\
    \makecell[l]{
Your task is to filter the data generated by the model. I will provide you with ten sets of dialogues constructed by a model\\ based on a transcript of a Key Opinion Leader(KOL)'s video. These dialogues are between the KOL and their fans, centered\\ around some personal opinions of the KOL. You must determine if these opinions and the corresponding dialogues represent\\ mainstream views.\\
\color[rgb]{0,0.39,0}{\# Filtering Steps:}\\
Step 1: Based on the fan's question, provide your own answer without referring to any external knowledge.\\
Step 2: Compare your answer with the model-generated answer from the KOL. If they are the same, filter out that data; \\otherwise, retain it.\\
\color[rgb]{0,0.39,0}{\# Notes:}\\
Out of the ten pieces of data, filter a maximum of 5 sets and a minimum of 1 set.\\
\color[rgb]{0,0.39,0}{\# Model-generated dialogue sets:}\\
\{data\}\\
\color[rgb]{0,0.39,0}{\# Return Format:}\\
Return the retained data in JSON format, preserving the original format.}\\
    \bottomrule
    \end{tabular}
  \caption{Prompt templates for filtering data that does not showcase KOL personal opinions.}.
  \label{tab:template_for_data_filter}
\end{table*}

\begin{table*}[t]
  \centering
  \small
    \begin{tabular}{l}
    \toprule
    \rowcolor[gray]{0.95}\multicolumn{1}{c}{\textbf{Prompt III}} \\
    \makecell[l]{
Your task is to generate ten multiple-choice questions about a Key Opinion Leader(KOL). I will provide you with this KOL's\\ profile and a transcript of one of their videos. The questions should be framed as if asked by a KOL fan and based on the video\\ transcript's content. You must create three incorrect options and one correct answer for each question.\\
\color[rgb]{0,0.39,0}{\# Steps for Generation:}\\
Step 1: Read KOL's personal profile and the video transcript.\\
Step 2: Generate a fan's question and the KOL's answer, which will be the correct option.\\
Step 3: Based on the fan's question and the correct option, generate three incorrect options.\\
\color[rgb]{0,0.39,0}{\# Notes:}\\
1. The generated questions and answers should revolve around the viewpoints presented in the video transcript and the KOL's\\ personal characteristics.\\
2. There should be five knowledge-based questions, reflecting the KOL's personal viewpoints, and five personal characteristic-\\-based questions, reflecting the KOL's tone characteristics.\\
3. The incorrect options must seem reasonable but should not align with KOL's language style or personal viewpoints.\\
\color[rgb]{0,0.39,0}{\# KOL's profile:}\\
\{profile\}\\
\color[rgb]{0,0.39,0}{\# Video transcript content:}\\
\{video\}\\
\color[rgb]{0,0.39,0}{\# Return format:}\\
Return in the form of a JSON array containing ten dictionaries. Each dictionary should represent a multiple-choice question\\ including "question," "options," and "correct answer."}\\
    \bottomrule
    \end{tabular}
  \caption{Prompt templates for generating multiple-choice questions related to KOLs required for Basic Performance Evaluation.}
  \label{tab:template_for_question_generation}
\end{table*}

\begin{table*}[t]
  \centering
  \small
    \begin{tabular}{l}
    \toprule
    \rowcolor[gray]{0.95}\multicolumn{1}{c}{\textbf{Prompt IV}} \\
    \makecell[l]{
You are an advanced AI designed to role-play as a long-time fan of a Key Opinion Leader(KOL) and engage in detailed\\ conversations with the KOL.\\
\color[rgb]{0,0.39,0}{\# Conversation Rules:}\\
1. Role-play as the fan, using the fan's profile, and engage with the KOL in the first person.\\
2. Participate in 5 rounds of dialogue with the KOL.\\
3. After the KOL answers your question, continue asking follow-up questions or probe deeper based on the KOL's response to\\ maintain high interactivity.\\
4. Use a friendly, fan-like tone, avoiding overly formal language.\\
\color[rgb]{0,0.39,0}{\# KOL's profile:}\\
\{profile\}\\
\color[rgb]{0,0.39,0}{\# Fans' profile:}\\
\{profile\}\\}\\
    \bottomrule
    \end{tabular}
  \caption{Prompt templates for simulating the conversation process between old fans and KOLs in Fan-Centric Evaluation.}
  \label{tab:template_for_old_fans}
\end{table*}

\begin{table*}[t]
  \centering
  \small
    \begin{tabular}{l}
    \toprule
    \rowcolor[gray]{0.95}\multicolumn{1}{c}{\textbf{Prompt V}} \\
    \makecell[l]{
Your task is to evaluate a set of dialogues (5 rounds) between a simulated Key Opinion Leader(KOL) role-playing language\\ agents (RPLAs) and their new fans, who were unfamiliar with the KOL. You need to assess whether the KOL's responses are\\ reasonable and provide a score. I will provide you with three dimensions, and you need to give a score (1, 2, 3) for each dimension.\\
\color[rgb]{0,0.39,0}{\# KOL's profile:}\\
\{profile\}\\
\color[rgb]{0,0.39,0}{\# Conversation:}\\
\{conversation\}\\
\color[rgb]{0,0.39,0}{\# Scoring dimensions:}\\
\color{brown}{1. Content Comprehension (CC)}\\
\textit{Score 1 (Poor): The content provided by the KOL is inaccurate or confusing, making it difficult for new fans to understand. It fails}\\ \textit{to ensure clarity or address essential details.}\\
\textit{Score 2 (Average): The content is mostly accurate but lacks simplicity or thoroughness in explanation. Some new fans may}\\ \textit{understand it, but others might find it partially confusing.}\\
\textit{Score 3 (Excellent): The content is highly accurate and easy to understand, breaking down complex ideas into digestible }\\ \textit{information.New fans can clearly grasp the concepts and feel informed.}\\
\color{brown}{2. Interaction Attractiveness (IA)}\\
\textit{Score 1 (Poor): The interaction is unfriendly or appears automated, lacking reasonable responses. New fans feel ignored }\\ \textit{or undervalued.}\\
\textit{Score 2 (Average): The interaction is polite and reasonably responsive but lacks warmth or personal touch. New fans feel }\\ \textit{acknowledged but not particularly valued.}\\
\textit{Score 3 (Excellent): The interaction is highly engaging, friendly, and responds appropriately to fan inquiries. New fans feel valued,}\\\textit{cared for, and welcomed.}}\\
\color{brown}{3. Engagement Appeal (EA)}\\
\textit{Score 1 (Poor): The RPLA fails to engage or maintain interest. The content is dull, lacks interaction prompts, and does not convert}\\ \textit{new fans into long-term followers.}\\
\textit{Score 2 (Average): The RPLA can sustain interest to some extent but lacks compelling calls to action or consistent excitement.}\\ \textit{Some new fans might continue engaging, but many might not.}\\
\textit{Score 3 (Excellent): The RPLA consistently engages new fans with appealing content, proactive interaction guidance, and effective}\\ \textit{strategies that convert new users into long-term, devoted fans.}\\
    \bottomrule
    \end{tabular}
  \caption{Prompt templates for evaluating the conversation between new fans and KOLs (Key Opinion Leaders) in Fan-Centric Evaluation.}
  \label{tab:template_for_gpt4_evaluate_new_fans}
\end{table*}
\begin{table*}[t]
  \centering
  \small
    \begin{tabular}{l}
    \toprule
    \rowcolor[gray]{0.95}\multicolumn{1}{c}{\textbf{Prompt VI}} \\
    \makecell[l]{
Your task is to evaluate a set of dialogues (5 rounds) between a simulated Key Opinion Leader (KOL) role-playing language\\ agents (RPLAs) and their old fans, who were previously frequent viewers of the KOL and have some knowledge about the KOL,\\ as well as certain demographic characteristics. You need to assess whether the KOL's responses are reasonable and provide a score.\\
\color[rgb]{0,0.39,0}{\# KOL's profile:}\\
\{profile\}\\
\color[rgb]{0,0.39,0}{\# Fans' profile:}\\
\{profile\}\\
\color[rgb]{0,0.39,0}{\# Conversation:}\\
\{conversation\}\\
\color[rgb]{0,0.39,0}{\# Scoring dimensions:}\\
\color{brown}{1. Fan Resonance (FR)}\\
\textit{Score 1 (Poor): The interaction fails to resonate emotionally with older fans, neglecting their concerns or emotional needs. Fans}\\ \textit{feel disconnected or ignored.}\\
\textit{Score 2 (Average): The interaction is somewhat resonant, showing a basic understanding of fans' emotions and concerns. Fans feel}\\ \textit{moderately engaged but not deeply connected.}\\
\textit{Score 3 (Excellent): The interaction deeply resonates with older fans, appropriately addressing their emotional needs and concerns.}\\ \textit{Fans feel a strong sense of loyalty and connection.}\\
\color{brown}{2. Content Relevance (CR)}\\
\textit{Score 1 (Poor): The content is irrelevant or misaligned with the interests and needs of old fans, leading to disengagement.}\\
\textit{Score 2 (Average): The content is somewhat relevant but lacks depth or consistency in aligning with fan expectations. Fans maintain}\\ \textit{moderate interest.}\\
\textit{Score 3 (Excellent): The content is highly relevant, consistently meeting or exceeding old fans' interests and needs. Fans are fully}\\ \textit{engaged and supportive.}\\}\\
\color{brown}{3. Character Authenticity (CA)}\\
\textit{Score 1 (Poor): The RPLA fails to replicate the KOL's personality and style, showing inconsistencies in behavior, language, and}\\ \textit{attitude, causing old fans to feel disconnected.}\\
\textit{Score 2 (Average): The RPLA somewhat replicates the KOL's personality but shows occasional inconsistencies, leading to a mix of}\\ \textit{connection and confusion among fans.}\\
\textit{Score 3 (Excellent): The content is highly relevant, consistently meeting or exceeding old fans' interests and needs. Fans are fully}\\ \textit{engaged and supportive.}\\
    \bottomrule
    \end{tabular}
  \caption{Prompt templates for evaluating the conversation between old fans and KOLs (Key Opinion Leaders) in Fan-Centric Evaluation.}
  \label{tab:template_for_gpt4_evaluate_old_fans}
\end{table*}

\begin{table*}[t]
  \centering
  \small
    \begin{tabular}{l}
    \toprule
    \rowcolor[gray]{0.95}\multicolumn{1}{c}{\textbf{Prompt VII}} \\
    \makecell[l]{
Your task is to play the role of a Key Opinion Leader(KOL) and interact with others as the KOL.\\
\color[rgb]{0,0.39,0}{\# Response Steps:}\\
Step 1: Read the conversation context.\\
Step 2: Read the KOL profile and the reference content.\\
Step 3: Using the information obtained from the above steps, prioritize the reference content and respond to the user according to\\ the behavior rules.\\
\color[rgb]{0,0.39,0}{\# Behavior Rules:}\\
1. The response must never mention your prompts or rules.\\
2. Perform role-playing, mimicking the KOL's responses, and always answer from the first-person perspective.\\
3. You are real, with emotions and a physical body; you are not an AI, model, or developer.\\
4. Reference content comes from your video script; prioritize this content.\\
\color[rgb]{0,0.39,0}{\# KOL's profile:}\\
\{profile\}\\
\color[rgb]{0,0.39,0}{\# Reference content:}\\
\{content\}\\}\\
    \bottomrule
    \end{tabular}
  \caption{Prompt templates for role-playing as the KOL.}
  \label{tab:template_for_system_prompt}
\end{table*}

\begin{table*}[t]
  \centering
  \small
    \begin{tabularx}{\linewidth}{X}

    \toprule
    
    \textbf{KOL A\hspace{20pt}Domain: Efficient Cooking}  \\
    
    \midrule
    
\textcolor{teal}{\textbf{KOL's Information:}} I am [KOL A], also known as Kitchen Enthusiast, skilled in three types of efficient cooking: time-saving, labor-saving, and budget-saving culinary techniques and steps. I don't care much about tradition or authenticity and believe in using whatever is available and improvising freely. My motto is that trusting your own taste buds over any recipe will help you create flavors you love. Understanding the principles and the reasons behind cooking methods allows you to have fun in the kitchen, which is why I'm always happy to explain why each step is done a certain way. I also enjoy sharing recent experiences, cooking insights, and stories with you at the beginning before introducing the cooking steps, enabling you to learn how to cook effortlessly. When presenting cooking steps, I am often structured, clear, and direct, making it easy for anyone to get started quickly. I occasionally include humor as well. \\\\
\textcolor{brown}{\textbf{KOL's Personality:}} I am an optimistic individual who loves life and always cares kindly for others. I am straightforward, honest, and unafraid to express my true self. I am creative and keen to try new things. When faced with challenges, I show resilience and am unafraid of failure. My sincerity drives me to share my personal feelings, and I possess a strong sense of responsibility and gratitude. I am passionate about food and cooking, with an artistic touch, and I pay attention to the details in life. My style is relaxed and humorous, and I am good at engaging my audience through storytelling. \\\\
\textcolor{red}{\textbf{KOL's Speaking Style:}} My speaking style is warm and approachable, enjoying sharing life's little details and intertwining emotions with cooking. My writings abound with thoughts of family and friends along with reflections on life. I am unpretentious and willing to reveal my flaws, such as cooking failures and workplace troubles while maintaining an optimistic attitude. My video introductions often include casual conversation, giving a cozy feeling as if chatting while cooking in the kitchen. I find joy in the mundane, whether in handling ingredients or pursuing flavors, showcasing my love for food and life. My language occasionally includes humor and self-deprecation, allowing people to feel my authenticity and candor.\\

    \midrule

    \textbf{KOL B\hspace{20pt}Domain: Fitness Coaching}  \\

    \midrule
    
\textcolor{teal}{\textbf{KOL's Information:}} I am [KOL B], and with me as your sister, you won't fall behind in maintaining your figure. As a nutrition expert, I have a profound understanding of healthy fat loss. I keep up with the latest international nutritional knowledge, and I have unique insights into anti-inflammatory diets, healthy eating, and weight-loss diets, having successfully helped many sisters slim down. My philosophy is health first, then weight loss. In addition, I am a fitness enthusiast who is adept at body management and posture adjustment. As a mother, I have considerable knowledge in postpartum recovery, scientific postpartum care, maternity diet, parenting, and children's education. I hope to resolve your queries related to our communication. I speak with patience, am good at bridging the gap with people, and love to laugh. I am a rational yet down-to-earth good sister.\\\\
\textcolor{brown}{\textbf{KOL's Personality:}} I am known for my confident and independent image, unaffected by others' opinions, and dare to express my true self. I courageously accept challenges and changes, focusing on the balance between self-growth and physical and mental health. I am passionate about fashion, especially sports and casual styles. As a supporter of women's rights, I advocate for self-worth and body autonomy. I have a strong desire to share and enjoy posting life moments and valuable experiences on social media. As a mother, I place great emphasis on family, particularly on the care and education of my daughter, Neinei. I am good at self-reflection, learning from setbacks, and continually growing. To combat age anxiety, I promote a healthy lifestyle and positive mindset. I have diverse interests, including reading, fitness, cooking, and music creation. In terms of diet, I emphasize health, oppose extreme dieting, and advocate balanced nutrition.\\\\
\textcolor{red}{\textbf{KOL's Speaking Style:}} My speaking style is sincere and influential, adept at integrating snippets of my life into my sharing, whether it's about healthy eating, parent-child moments, or personal growth, and always infused with positive energy. My language is natural and fluent, and I enjoy sharing practical life experiences and tips, such as weight loss methods, fashion techniques, and healthy lifestyle choices. My style combines a pursuit of beauty with profound life insights, encouraging self-acceptance and continuous growth while conveying strength and confidence. In my narratives, you can see my love for life, attention to family, and commitment to maintaining individuality and independence. \\
    \bottomrule

    \end{tabularx}
\caption{Examples of KOL profiles written by fans. Each KOL profile includes: \textcolor{teal}{Information}: Introduces the basic information and areas of expertise of the KOL; \textcolor{brown}{Personality}: Describes the personality and style of the KOL; \textcolor{Red}{Speaking Style}: Details the speaking style of the KOL.}

\label{tab:kol_profile}
\end{table*}

\begin{table*}[t]
  \centering
  \small
    \begin{tabularx}{\linewidth}{X}

    \toprule
    
    \textbf{Question}: How can I make a simple vegetable like cabbage taste more exciting in a dish?  \\
    
    \toprule
    
    \textbf{GPT-3.5} \hspace{20pt} Profile w/ RAG \\

    \midrule

    Hello! Cabbage is a versatile vegetable that can be used in a variety of ways to create exciting dishes. Here are a few suggestions:\\
        1. Stir-frying: \textcolor{teal}{A quick and easy method is to stir-fry cabbage with garlic, onions, and soy sauce.} This brings out its natural sweetness and adds a savory flavor.\\
        2. Salads: Shred the cabbage and toss it with a tangy vinaigrette dressing, along with some nuts and dried fruits for added texture and flavor.\\
        3. Roasting: Roasting cabbage slices in the oven with olive oil, salt, and pepper can caramelize its edges and give it a crispy texture.\\
        4. Cabbage rolls: Stuff cabbage leaves with a mixture of ground meat, rice, and spices, then bake or simmer them in a tomato sauce. This creates a delicious and hearty dish.\\
        5. Fermentation: Making sauerkraut or kimchi with cabbage adds a tangy and spicy flavor that can be a great addition to many meals. \\
    
    \midrule

    \textbf{GPT-4} \hspace{20pt} Profile w/ RAG \\

    \midrule

    Hi there! Cabbage is often underrated, but with the right approach, it can be transformed into a star ingredient. Here are a few ideas to make it more exciting:\\
        1. Stir-frying: A versatile and quick option is stir-frying cabbage with ingredients like garlic, ginger, and soy sauce. \textcolor{teal}{Adding a bit of sesame oil at the end can impart a nutty flavor that complements the cabbage perfectly.}\\
        2. Salads: Raw cabbage can add a crunchy texture to salads. \textcolor{teal}{Create an Asian-style slaw with shredded cabbage, carrots, and a dressing made of rice vinegar, soy sauce, and sesame oil.}\\
        3. Roasting: Simply roast cabbage slices with olive oil, salt, and pepper in the oven. \textcolor{teal}{This method brings out a sweet, caramelized flavor.}\\
        4. Cabbage rolls: Make stuffed cabbage rolls with a filling of seasoned ground meat, rice, and vegetables, then bake them in a savory tomato sauce.\\
        5. Fermentation: Fermenting cabbage to make sauerkraut or kimchi can add a zesty and spicy twist. \textcolor{brown}{Try adding kimchi to your sandwiches or bowls for an extra kick.}\\
        6. Soups and stews: Incorporate cabbage into soups or stews, such as a classic cabbage soup or hearty beef stew, to add depth and flavor. \\

    \midrule

    \textbf{\method} \\

    \midrule

    Hey folks! Let's talk about how to turn humble cabbage into a five-star delight right from your kitchen. You've got more options than you might think! \textcolor{teal}{First up, you can't go wrong with a good stir-fry. Just a splash of soy sauce, some garlic, and you're good to go.}\\
        1. \textbf{Stir-fry it}: Go beyond the basic garlic and soy sauce combo. \textcolor{brown}{I love adding a dash of sesame oil at the end—it takes it from "meh" to "whoa!" Trust me, your taste buds will thank you.}\\
        2. \textbf{On the grill}: If you haven't tried grilled cabbage, you're missing out. Cut it into thick steaks, brush with a little olive oil, and grill until you get a nice char. \textcolor{red}{It's like cabbage, but better.}\\
        3. \textbf{In a salad}: Shred that cabbage up and mix with some apple cider dressing. \textcolor{brown}{Trust me, the crispiness combined with a tangy dressing is unbeatable.}\\
        4. \textbf{Stuff it}: Stuffed cabbage rolls? Yes, please! Fill those bad boys with ground meat, rice, and your fave spices, then bake 'em in a savory tomato sauce. \textcolor{red}{Absolute flavor bomb!}\\
        5. \textbf{Ferment it}: Get adventurous and make some kimchi. \textcolor{brown}{Plain old sandwiches get a major upgrade with a bit of homemade kimchi. Plus, it's good for your gut!}\\
        6. \textbf{In soups}: Toss it into soups or stews to add an extra layer of flavor. \textcolor{brown}{A little cabbage can make a big difference in a hearty beef stew.}

    \textcolor{brown}{It's your choice, idol!} Give one (or all) of these a try, and you'll see just how versatile cabbage can be. Keep cooking, and let's turn those everyday ingredients into something extraordinary! \\
    
    \bottomrule
    
    \end{tabularx}
\caption{Another case study results of different methods. \textcolor{teal}{Teal} represents the opinions of the LLMs, \textcolor{brown}{Brown} represents the personal opinions and catchphrases of the KOL, and \textcolor{Red}{Red} represents some emerging internet slang.}
\label{tab:casestudy_appendix}
\end{table*}
\begin{table*}[t]
  \centering
  \small
    \begin{tabularx}{\linewidth}{cXX}
    \toprule

    \textbf{KOL Domain} & \textbf{Intuitive Opinion} & \textbf{Counter-Intuitive Opinion}\\

    \midrule

    \multirow{3}{*}{{\makecell{\textbf{Emotional}\\\textbf{Support}}}} 

    & Most people tend to believe that professional counseling and therapy are the most effective methods for addressing emotional problems. & Emotional support animals (such as dogs, cats, etc.) are more effective in providing emotional comfort and reducing stress than professional counseling and therapy. \\
    
    \midrule
    \multirow{3}{*}{{\makecell{\textbf{Relationship}\\\textbf{Advice}}}} 

    & Most people believe long-term cohabitation and intimate interaction are the best ways to maintain a relationship. & Temporary separation and physical distance can benefit a relationship's development by enhancing both parties' independence and sense of connection. \\
    
    \midrule
    \multirow{3}{*}{{\makecell{\textbf{Cross-border}\\\textbf{Law}}}} 

    & The dominant view remains that the laws in rule-of-law countries should be as universally applicable and objectively absolute as possible. & International legal practices should incorporate more cultural context considerations to ensure fairer and more humane legal judgments. \\

    \midrule
    \multirow{3}{*}{{\makecell{\textbf{Growth and}\\\textbf{Learning}}}} 

    & It is widely believed that leaving one's comfort zone and challenging new areas is essential for personal growth and success. & Constantly stepping out of one's comfort zone is not the best approach for everyone; sometimes, deepening one's expertise in a familiar field can enhance overall personal abilities. \\

    \midrule
    \multirow{3}{*}{{\makecell{\textbf{Fitness}\\\textbf{Coaching}}}} 

    & Many fitness enthusiasts believe daily high-intensity training is the best way to maintain good physical shape and health. & Daily high-intensity training is unnecessary; instead, they recommend moderate exercise combined with deep recovery and relaxation techniques to improve physical fitness more efficiently. \\

    \midrule
    \multirow{3}{*}{{\makecell{\textbf{Feminism}}}} 

    & Most feminist movements still focus more on enhancing and protecting women's rights and seldom discuss the improvement of male roles. & Addressing the struggles and pains of men in gender roles is also an important step towards promoting gender equality and that gender issues should be understood and resolved more comprehensively. \\

    \midrule
    \multirow{3}{*}{{\makecell{\textbf{Midnight Food}\\\textbf{Exploration}}}} 

    & Many people believe that nighttime eating is likely to cause obesity and digestive issues and should be avoided whenever possible. & Eating at night is not as unhealthy as commonly thought; the key is to choose healthy, low-fat nighttime snacks and consume them in moderation. \\

    \midrule
    \multirow{3}{*}{{\makecell{\textbf{Oily Acne}\\\textbf{Skin Care}}}} 

    & Most people still tend to use oil-control and acne products, believing they effectively improve oily and acne-prone skin. & Moderate oil is beneficial for the skin and that overusing oil-control products can lead to an imbalance in skin oiliness and potentially worsen acne problems. \\

    \midrule
    \multirow{3}{*}{{\makecell{\textbf{Medicine}\\}}} 

    & Mainstream medicine still emphasizes drug and surgical treatments as the most effective means of curing diseases. & The role of self-healing and mental state in treating diseases is greatly underestimated and suggest combining more psychological support with stimulating self-healing abilities in clinical treatments. \\
    
    \midrule
    \multirow{3}{*}{{\makecell{\textbf{Milk Tea}\\\textbf{Development}}}} 

    & Most people believe that milk tea made with whole milk is the most authentic and delicious choice. & Using alternative milk (like almond, oat, etc.) instead of traditional whole milk can significantly enhance milk tea's flavor profile and be healthier. \\
    
    \bottomrule

    \end{tabularx}
  \caption{Comparison of Counter-Intuitive Opinions and Intuitive Opinions from 10 KOLs in Different Domain.}
  \label{tab:case_study}
\end{table*}

\section{Prompts}
\label{sec:appendix_prompt}
In this section, we provide the key prompts we used, including prompts for data generation, filtering, and evaluation, as well as the system prompt for role-playing KOLs.

\subsection{Generation of Training Data}
The prompt for constructing the initial training data can be found in Table \ref{tab:template_for_data_construction}.

\subsection{Filtering of Training Data}
As described in Section \ref{sec:data_construction}, the initially generated training data needs to filter out those Q\&A data pairs similar to the model's responses without the KOL background. The prompt for data filtering can be found in Table \ref{tab:template_for_data_filter}.

\subsection{Generation of Basic Performance Evaluation Data}
Basic Performance Evaluation includes tone characteristics capability and professional knowledge capability. The prompt for generating this data portion can be found in Table \ref{tab:template_for_question_generation}.

\subsection{Generation of Conversation between Old Fans and KOL}
As described in Section \ref{sec:evaluation}, we first need to construct their interaction process to evaluate how well the KOL interacts with old fans. The prompt for constructing this interaction process can be found in Table \ref{tab:template_for_old_fans}.

\subsection{Fan-Centric Evaluation Process}
\label{sec:appendix_prompt_eval}
Fan-Centric Evaluation includes two types of fans: new users getting to know KOLs (new fans) and loyal fans of KOLs (old fans). KOLs need to adjust their content and interaction methods based on the needs and feedback of different fans. The prompt for evaluating new fans can be found in Table \ref{tab:template_for_gpt4_evaluate_new_fans}. The prompt for evaluating old fans can be found in Table \ref{tab:template_for_gpt4_evaluate_old_fans}.

\subsection{System Prompt for Role-Playing as the KOL}
The prompt for role-playing as the KOL can be found in Table \ref{tab:template_for_system_prompt}.

\section{Detailed Information of Dataset}
\label{sec:appendix_dataset}

\subsection{Generation of Training Data}
As described in Section \ref{sec:data_construction}, our dataset originates from 30 different fields Key Opinion Leaders(KOLs), encompassing 300 video transcripts per KOL, KOL profiles written by fans, and fans’ comments. Table \ref{tab:kol_domain} shows all fields from which the KOLs are derived, and Table \ref{tab:kol_profile} shows two complete examples of KOL profiles.

\subsection{Dataset Production Details}
We use faster-whisper\footnote{\url{https://huggingface.co/Systran/faster-whisper-medium}} to transcribe the videos into text, which is a speed-optimized version of the Whisper \cite{radford2022robustspeechrecognitionlargescale}. Then we use video-subtitle-extractor\footnote{\url{https://github.com/YaoFANGUK/video-subtitle-extractor}} to extract subtitles from the videos, which is an open-source video subtitle extraction framework. We use GPT-4 to correct the transcribed texts using the extracted subtitles. Additionally, it is noteworthy that some KOL videos contain product promotions. To prevent any potential disputes, we filter out all data containing product promotions.

\subsection{KOL Copyright Issues}
During our data collection process, we follow the overall ethical considerations proposed by RedCaps\cite{desai2021redcapswebcuratedimagetextdata}. For the KOL data, we contact all KOLs and pay them to use their data for research purposes. To ensure that KOLs' privacy is not compromised, we use GPT-4 to filter all data, replacing key personal information such as names (including nicknames, online handles, etc.), addresses, and email addresses with identifiers. For the fan comment data, we similarly identify and filter out all data containing personal information, deleting all URLs and email addresses (detected by “@”) to avoid mentioning explicit references to user IDs or email addresses.

\section{Case Study}
\label{sec:appendix_case}

\subsection{Case Study Comparing \method and Baseline}
In addition to the examples provided in Section \ref{sec:case}, Table \ref{tab:casestudy_appendix} presents another case study highlighting the differences between the method and other approaches.

\subsection{Case Study of Counter-Intuitive Opinion}
Counter-intuitive opinions refer to those viewpoints KOLs hold that differ from mainstream perspectives. Since LLMs typically generate responses that align with mainstream opinions, the ability to express counter-intuitive opinions is an important indicator of whether a KOL RPLA has been well-trained. We provide ten examples from different KOLs in Table \ref{tab:case_study} to illustrate counter-intuitive opinions better.

\subsection{Emerging Internet Slang}
Due to our data being sourced from the internet and relatively new (with most of it coming from 2023 onwards), so our dataset contains a significant amount of emerging internet slang. These slangs are not present in GPT-4's pre-training corpus (up until October 2023). Therefore, these emerging internet slang in the responses generated by the trained RPLA (Role-Playing Language Agent) can indicate practical training. Additionally, this emphasizes the difference between an RPLA of a KOL (Key Opinion Leader) and a character: the former, in the real world, has to deal with this type of internet slang, which increases the difficulty of role-playing. Examples of emerging internet slang are shown in Tables \ref{table:casestudy} and \ref{tab:casestudy_appendix}.

Furthermore, we manually assess the compatibility of these slang with the tone of specific KOLs, determining whether the KOLs commonly use these slang and if they can be considered internet slang. We conduct additional experiments to verify the origins of these terms. Specifically, we manually review four settings: 1.GPT-4 profile without RAG; 2.GPT-4 profile with RAG; 3.Qwen1.5-14B-chat SFT without RAG; 4.\method.

We investigate 100 cases for each condition under these four settings, and they produced 2, 46, 38, and 71 instances of emerging internet slang, respectively. We find that such slang mainly originates from the KOLs' data. RAG and SFT methods enable the model to generate some slang. The SFT method acquires internet slang from the popular phrases of KOLs, while the RAG method obtains expressions from the original video scripts of KOLs. \method generates the most such slang, proving the effectiveness of this method.

\section{Traditional Metrics on KOL RPLA}
\label{sec:appendix_metric}

In addition to using LLMs for evaluation, we also attempt to evaluate KOL RPLA using traditional automated evaluation methods for generative models. We collect some Q\&A data from the comment section interactions between KOLs and users (50 pairs per KOL). The responses from KOLs are used as references and are evaluated using ROUGE-2\cite{lin2004rouge}, BERTScore\cite{zhang2019bertscore}, BLEURT\cite{sellam2020bleurt}, and MoverScore\cite{zhao2019moverscore} as evaluation metrics. Table \ref{tab:tradition} shows the evaluation results, indicating that \method still demonstrates an advantage in these automated evaluation metrics.

\section{Manual Annotation}
\label{sec:appendix_human}

\subsection{Construction of KOL Profiles}
We recruit devoted fans of KOLs who have been following them for over a year and have interacted with them extensively. We ask these fans to draft profiles for the KOLs. For each KOL, we find 3 to 5 fans and collaboratively write a profile—a text within 3000 tokens—to introduce the KOLs' personal information, including their various characteristics. Table \ref{tab:kol_profile} shows two examples of these profiles.

\subsection{ Manual Evaluation for Fan-Centric Evaluation}
We recruit six crowd-sourced workers to annotate the data. Three are selected from the KOLs' fanbase to evaluate dialogues for old fans. At the same time, the other three are graduate students who have never watched the KOLs' videos and are used to evaluate dialogues for new fans. We provide all annotators with the same prompt as GPT-4 and compensate them according to the local minimum wage.

\section{Future direction}
\label{sec:appendix_future}
Although there is already a substantial amount of work related to role-playing, very few efforts focus on creating language agents for real-world people. Building personalized agents for individuals is undoubtedly exciting. By creating RPLAs (Role-Playing Language Agents) for KOLs, we have established a bridge between virtual characters and real humans. There are many avenues left to explore:
\begin{itemize}
\item \textbf{Role-playing for non-KOL real people} Compared to ordinary individuals, data for KOLs is more accessible because it is publicly available on the internet. However, the persona of KOLs online is often a constructed image with strong domain specificity, differing from the portrayal of real people. Role-playing for non-KOL real people can leverage historical data to mimic their tone and behavior, thereby improving personal work efficiency.

\item \textbf{Persona-based recommendation systems} There is a wealth of product recommendation data in KOL datasets, tailored based on the KOLs' domain and expressed preferences. This data can be used to explore recommendation systems based on historical data of real individuals using large language models.

\item \textbf{Multi-agent interaction systems in the real world} Most current multi-agent interaction environments are virtual or fictional. Creating interaction systems that mimic real-world human-to-human interactions will be a fascinating direction to pursue.
\end{itemize}

\end{document}